\documentclass[11pt]{article}

\usepackage[preprint]{acl}

\usepackage{times}

\usepackage{latexsym}

\usepackage[T1]{fontenc}

\usepackage{tabularx}
\usepackage{array}
\usepackage{multirow}
\usepackage{float}

\usepackage[utf8]{inputenc}

\usepackage[protrusion=false]{microtype}

\usepackage{inconsolata}

\usepackage{graphicx}

\usepackage{amsmath}

\usepackage{amssymb}

\usepackage{booktabs}

\usepackage{multirow}

\usepackage{colortbl}

\usepackage[most]{tcolorbox}
\usepackage{tabularx}
\usepackage{array}
\usepackage{float}

\usepackage[section]{placeins}

\definecolor{takeawaynavy}{RGB}{31,43,69}
\definecolor{takeawaybg}{RGB}{248,249,252}
\definecolor{takeawayborder}{RGB}{132,143,164}

\newtcolorbox{takeawaypanel}[1][Takeaway]{%
  enhanced,
  colback=takeawaybg,
  colframe=takeawayborder,
  boxrule=0.5pt,
  arc=1pt,
  left=8pt,
  right=8pt,
  top=9pt,
  bottom=5pt,
  before skip=0.8\baselineskip,
  after skip=0.7\baselineskip,
  title=#1,
  coltitle=white,
  fonttitle=\bfseries,
  fontupper=\normalfont,
  attach boxed title to top left={xshift=6pt,yshift=-2.5mm},
  boxed title style={%
    colback=takeawaynavy,
    colframe=takeawaynavy,
    boxrule=0pt,
    arc=1pt,
    left=5pt,
    right=5pt,
    top=2pt,
    bottom=2pt
  }
}
\newcommand{\takeawaybox}[1]{\begin{takeawaypanel}#1\end{takeawaypanel}}

\title{Depth as a Candidate Source: Portfolio Allocation for Recurrent Test-Time Scaling in Large Language Models}

\author{%
  \textbf{Chaoyuan Hao, Wentao Yue, Yanyan Wu, Boyuan Gao, Tianyou Lai} \\
  \textbf{Maolin Wang, Yongxin Ni, Hanwen Du, Youhua Li}
}

\hypersetup{%
  pdftitle={Depth as a Candidate Source: Portfolio Allocation for Recurrent Test-Time Scaling in Large Language Models},
  pdfauthor={Chaoyuan Hao, Wentao Yue, Yanyan Wu, Boyuan Gao, Tianyou Lai, Maolin Wang, Yongxin Ni, Hanwen Du, Youhua Li}
}

\begin{document}

\maketitle

\begin{abstract}
Recurrent latent reasoning in large lanugage models (LLMs) scales test-time computation by increasing recurrent depth, naturally framing inference as a problem of deciding when to stop. We identify a previously overlooked limitation of this single-depth view through systematic preliminary analysis: deeper recurrence can lose correct solutions found at shallower depths, while depth-specific success sets remain broadly non-nested even after controlling for marginal accuracy. This suggests that depth determines not only computation cost, but also access to distinct candidate solutions. Motivated by this finding, we propose \emph{Depth Portfolio Allocation} (DPA), which reformulates recurrent test-time scaling as budgeted allocation across depth-specific candidate sources. DPA supports both fixed portfolios and feedback-guided sequential allocation under a shared inference budget. Experiments on GSM8K, SVAMP, MultiArith, and MBPP show that mixed-depth portfolios expand oracle coverage under matched candidate counts and nominal costs, while feedback-guided allocation generally matches or improves uniform allocation and can reduce computation in high-budget settings. Code and data are avaible at : \textcolor{blue}{\url{https://anonymous.4open.science/r/LLM-latent-B7BC/}}.
\end{abstract}

\raggedbottom
\widowpenalty=10000
\clubpenalty=10000
\displaywidowpenalty=10000
\emergencystretch=0.6em

\section{Introduction}
\label{sec:introduction}

Test-time scaling (TTS) in LLMs improves performance by allocating additional inference-time computation without changing model parameters \citep{snell2025scaling}. Existing approaches typically spend this compute in token space, for example through longer reasoning traces \citep{wei2022chain}, multiple candidate samples \citep{wang2023selfconsistency}, or verifier-based selection \citep{cobbe2021training}.  Recurrent latent reasoning provides a different form of TTS: instead of generating longer explicit traces, the model repeatedly applies recurrent computation to its hidden representations, allowing answers to be decoded after different numbers of recurrent updates \citep{geiping2025scaling}. The recurrent depth therefore directly controls how much latent computation is spent on a question, making depth selection a natural inference-time decision.

A common view of recurrent test-time scaling is consequently to ask how much recurrent computation to use, or when computation should stop \citep{geiping2025scaling,graves2016adaptive,elbayad2020depth}. This single-trajectory formulation is most natural when additional recurrence progressively refines, or at least preserves, solutions available at shallower depths. Recent work on looped language models, however, shows that performance can peak and then degrade as recurrence continues \citep{yang2026stars}, suggesting that deeper computation need not provide a monotonic refinement. If so, selecting a single stopping depth may overlook useful at other depths solutions that emerge elsewhere along the recurrent trajectory.

We examine this issue at the level of individual questions and candidate solutions through the preliminary analysis in Section~\ref{sec:understanding-steps}. Across a large set of recurrent reasoning trajectories, different questions favor different depths, deeper recurrence does not always preserve earlier correct answers, and depth-specific success sets remain broadly non-nested even after controlling for marginal accuracy. These findings suggest that depth determines not only computational cost, but also access to partially distinct candidate solutions. Recurrent inference is therefore not only a question of \emph{where to stop}, but also of \emph{how to allocate a finite budget across depth-specific candidate sources}.

\begin{figure*}[!t]
    \centering
    \includegraphics[width=\textwidth]{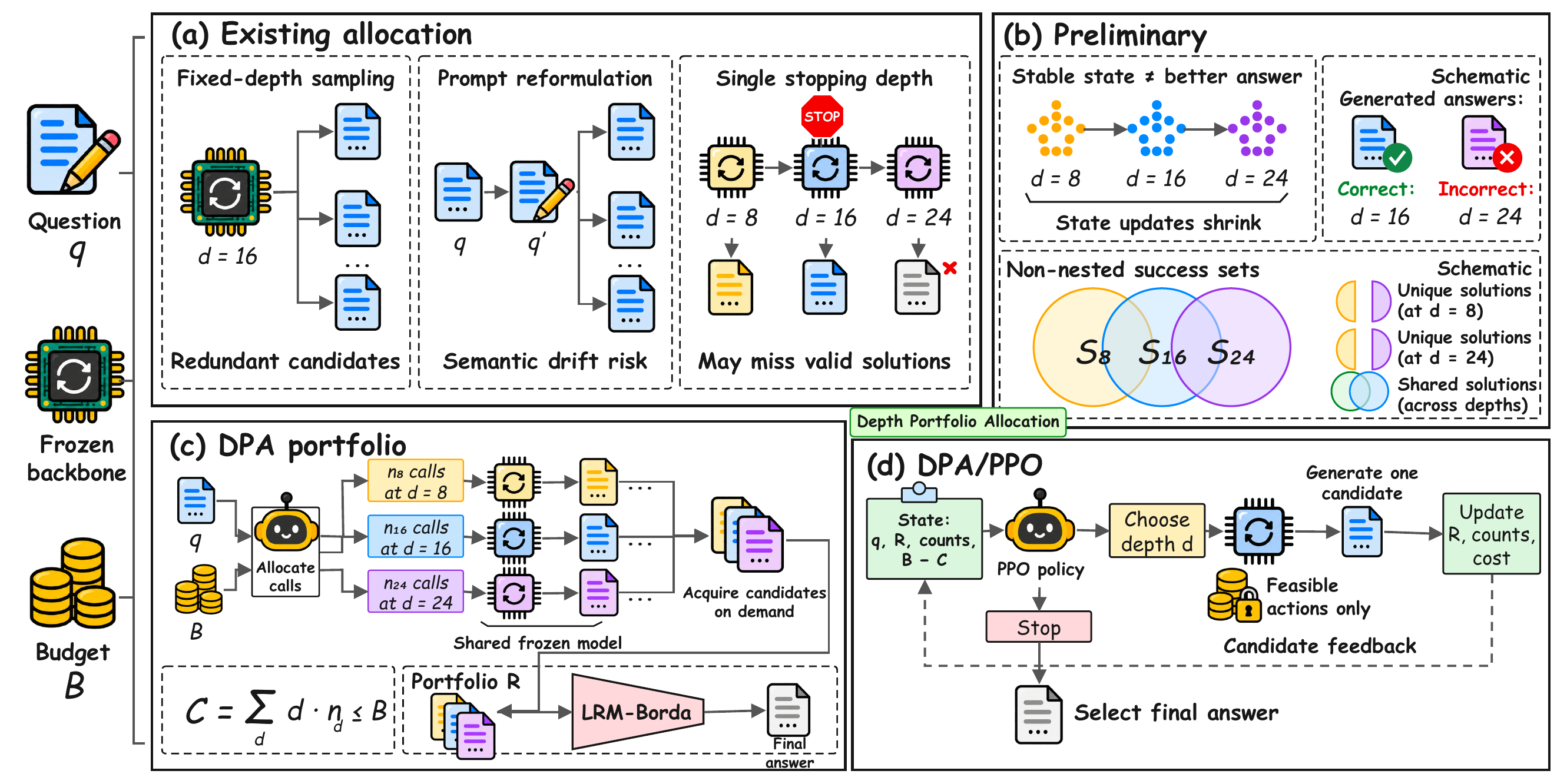}
    \caption{\textbf{Motivation and framework.} (a) Existing single-depth and sampling strategies; (b) preliminary evidence that deeper recurrence does not subsume shallower solutions; (c) DPA allocates a shared budget across depth-specific candidate sources; (d) feedback-guided allocation chooses a depth or stops. States and sets are schematic.}
    \label{fig:motivation}
\end{figure*}

\begin{takeawaypanel}[Research Question]
Does a single stopping depth miss correct solutions available at other depths, and how can a limited budget exploit these complementary sources?
\end{takeawaypanel}

Motivated by this preliminary diagnosis, we propose \emph{Depth Portfolio Allocation} (DPA), which treats recurrent depths as candidate sources with different nominal costs and allocates acquisition counts under a shared budget (Figure~\ref{fig:motivation}). DPA separates the allocation problem from its optimizer: portfolios may be fixed in advance or chosen sequentially from candidate feedback. In our sequential implementation, a standard PPO policy observes the current candidate set and remaining budget, then decides which depth to query next or whether to stop, allowing the allocation to adapt to the candidates acquired so far. Experiments across multiple datasets (e.g., GSM8K, SVAMP, MultiArith, and MBPP) consistently validate the proposed allocation framework in both solution coverage and budget efficiency.

Our contributions are threefold:
\begin{itemize}

    \item We identify and empirically establish a previously overlooked issue in single-depth recurrent scaling: deeper recurrence can lose correct shallower-depth solutions, while depth-specific success sets remain systematically non-nested, making a single stopping depth incomplete.

    \item A new formulation of recurrent test-time scaling as budgeted allocation across depths. Depth Portfolio Allocation (DPA) treats each depth as both a computation cost and a candidate source, and allocates a shared inference budget across multiple depths rather than selecting a single stopping point.

    \item Extensive evaluation on four datasets. Mixed-depth portfolios consistently expand oracle coverage under matched candidate and nominal-cost budgets, while feedback-guided allocation generally matches or improves uniform allocation and can leave part of the available computation budget unused without sacrificing accuracy.
\end{itemize}

\section{Related Work}
\label{sec:related-work}

\paragraph{Test-time scaling and candidate selection.}
Test-time scaling spends additional inference compute on explicit reasoning, repeated sampling, search, or verification
\citep{wei2022chain,wang2023selfconsistency,yao2023tree,cobbe2021training,lightman2024verify,zhang2025generative}.
Recent work further studies Best-of-$N$, compute-optimal inference, and diversity-aware sampling, showing that the value of additional compute depends on candidate coverage, problem difficulty, and verification quality
\citep{snell2025scaling,wu2025inference,setlur2025scaling,huang2025bestofn,wu2026mode}.
DPA studies a complementary allocation problem: how a fixed budget should be distributed across candidate sources induced by recurrent depth.

\paragraph{Adaptive computation.}
Adaptive-computation methods learn when to halt or how much depth to allocate, from Adaptive Computation Time and Universal Transformers to PonderNet, depth-adaptive Transformers, and early-exit language models
\citep{graves2016adaptive,dehghani2019universal,banino2021pondernet,elbayad2020depth,zhou2020patience,schuster2022calm}.
These methods primarily determine a single amount of computation per input; our setting instead allows multiple depths to contribute candidates under one shared budget.

\paragraph{Recurrent and continuous latent reasoning.}
Latent reasoning moves computation from explicit token traces into hidden-state refinement through pause tokens, continuous thoughts, and looped or recurrent Transformers
\citep{goyal2024pause,hao2025continuous,saunshi2025reasoning,geiping2025scaling}.
Recent work further explores learned or elastic recurrent depth, recurrent stability, and efficient sampling
\citep{zhu2025ouro,jeddi2026loopformer,yang2026stars,geiping2026parallel}.
Hidden states can also encode evaluative or reward signals
\citep{wang2025coe,du2026latent}.
We instead ask whether different recurrent depths expose complementary, non-nested candidate sets and how a shared budget should be allocated across them; PPO serves only as a standard optimizer for the sequential policy 
\citep{schulman2017ppo}. A concurrent study \citep{anonymous2026agreement} evaluates
probability updating under copied and independent evidence
using exact Bayesian references. DPA instead studies candidate
complementarity across recurrent depths and budgeted inference
allocation.

\section{Preliminary: Diagnosing Single-Depth Recurrent Scaling}

\begin{figure}[!t]
    \centering
    \includegraphics[width=\linewidth]{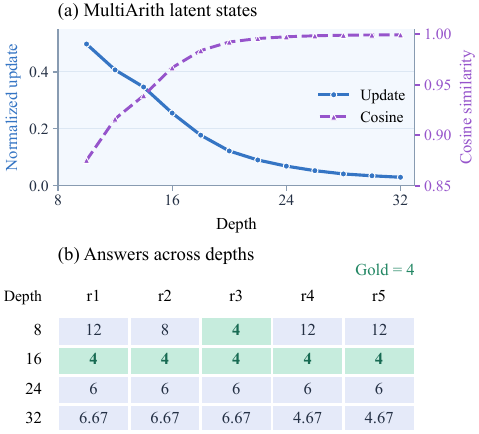}
    \caption{\textbf{Stability does not ensure correctness.} (a) Latent updates and cosine similarity; (b) five answers per depth, with correct answers in green.}
    \label{fig:latent-convergence}
    \vspace{-3mm}
\end{figure}

\label{sec:understanding-steps}

\subsection{Deeper Recurrence Does Not Preserve Correctness}

A single-depth view implicitly assumes that additional recurrence progressively refines the current solution. We first compare latent-state changes with answers decoded at different recurrent depths on Huginn (Figure~\ref{fig:latent-convergence}).

\textbf{Latent updates gradually diminish with recurrent depth.} Figure~\ref{fig:latent-convergence}(a) shows decreasing normalized update magnitudes on MultiArith and adjacent-state cosine similarities approaching one, an empirical trend toward more stable latent states.

\textbf{Stable representations do not ensure that deeper decoding retains correct answers from shallower depths.} The example in Figure~\ref{fig:latent-convergence}(b) asks how many shelves are needed for $14-2$ remaining books, with three books per shelf. The correct answer, 4, appears in samples at $d=8,16$ but not at $d=24,32$. Thus, additional recurrence need not improve answer quality monotonically, and deeper stopping can miss earlier correct solutions.

\textbf{The preliminary observation motivates studying depth complementarity.} Because complete answers can be decoded from different intermediate states, a depth can act as a distinct candidate source rather than only a stopping position. We ask whether several depths cover questions missed by a single depth before turning to budget allocation.

\subsection{Correct Solutions Are Complementary across Depths}

\label{sec:complementary-depth-solutions}

With inputs fixed, we compare answer diversity and success sets over $\mathcal{D}=\{8,10,\ldots,32\}$, examining depth preferences and whether deeper success sets subsume shallower ones.

\begin{figure}[!htbp]

    \centering

    \includegraphics[width=\linewidth]{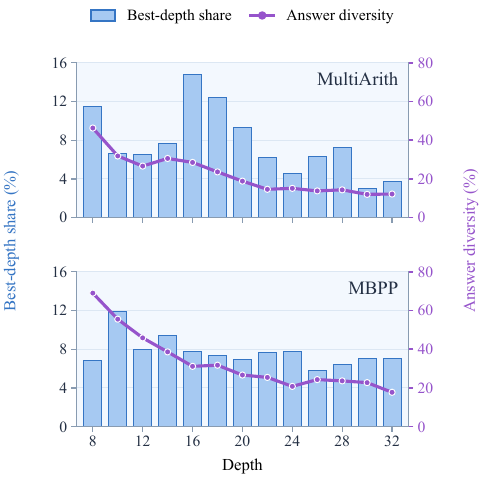}

\caption{\textbf{Depth preferences and answer diversity.} Bars show best-depth frequencies; curves show disagreement across repeats at each depth.}

    \label{fig:step-preference-diversity}

\end{figure}

\textbf{Answer diversity decreases with depth, but preferred depths remain dispersed.} In Figure~\ref{fig:step-preference-diversity}, bars indicate the fraction of questions for which each depth is best, and curves show within-depth answer diversity. On MultiArith and MBPP, answer diversity generally declines with depth, while best-depth preferences span multiple positions. Candidates become more similar overall, yet different questions still favor different depths, indicating that answers from multiple depths can be useful.

\paragraph{Overlap between success sets.}

We sample five answers per question at each of the 13 depths $\mathcal{D}=\{8,10,\ldots,32\}$ on MultiArith and MBPP. Let $y_{qdr}\in\{0,1\}$ indicate whether repeat $r$ for question $q$ at depth $d$ is correct. The success set at depth $d$ is

\begin{equation}
C_d=\{q:\max_{r\in\{1,\ldots,5\}} y_{qdr}=1\}.
\end{equation}

\begin{figure}[!t]

    \centering

    \includegraphics[width=\linewidth]{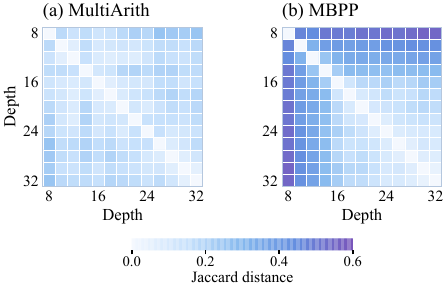}

\caption{\textbf{Jaccard distances between depth-specific success sets.} Both tasks share a color scale; diagonal entries are zero.}

    \label{fig:success-set-jaccard}

\vspace{-3mm}
\end{figure}

Two success sets are \emph{mutually non-nested} if each contains questions absent from the other. We measure their difference using Jaccard distance, $J(d_i,d_j)=1-|C_{d_i}\cap C_{d_j}|/|C_{d_i}\cup C_{d_j}|$. Both datasets use the same depth grid; Figure~\ref{fig:success-set-jaccard} reports distances for all depth pairs.

\textbf{Success sets at different depths are broadly mutually non-nested.} Adjacent depths typically have more similar success sets, whereas more distant depths differ more. Table~\ref{tab:success-set-summary} shows that nearly every pair on both tasks contains questions solved by only one of its depths. Deeper success sets therefore do not simply subsume shallower ones: combining depths can contribute different correct solutions to a candidate portfolio.

\begin{table}[!t]

\centering

\scriptsize

\setlength{\tabcolsep}{3pt}

\caption{\textbf{Mutually non-nested pairs and depth association after controlling for marginal accuracy.}}

\label{tab:success-set-summary}

\resizebox{\columnwidth}{!}{%

\begin{tabular}
{lrl}
\toprule
Dataset & Non-nested pairs & $\Delta J$ per 8 steps (95\% CI) \\
\midrule
MultiArith & 77/78 & +3.36 pp [0.87, 5.83] \\
MBPP & 78/78 & +5.63 pp [3.48, 8.53] \\
\bottomrule
\end{tabular}%

}

\end{table}

To examine whether the association between depth separation and success-set differences persists after accounting for marginal accuracy, we regress pairwise Jaccard distance on depth separation, the absolute difference in single-sample marginal accuracy, and mean marginal accuracy:

\begin{equation}
\begin{aligned}
J(d_i,d_j)={}&\beta_0+\beta_1|d_i-d_j|\\
&+\beta_2|a_i-a_j|+\beta_3\frac{a_i+a_j}{2}+\epsilon_{ij}.
\end{aligned}
\end{equation}

Here, $a_i$ and $a_j$ are the marginal accuracies at the two depths. We bootstrap questions 5,000 times, recomputing distances and regression coefficients in each replicate. Table~\ref{tab:success-set-summary} summarizes the depth-separation association after controlling for marginal accuracy.

To test whether complementarity is explained solely by question difficulty and marginal accuracy, we perform 5,000 fixed-margin randomizations of the observed correctness matrix, preserving each question's total successes and each depth--repeat slot's success count.

\textbf{Depth-dependent success-set differences exceed the fixed-margin null expectation.} On both tasks, the association between depth separation and success-set differences is significantly greater than expected under this fixed-margin null, supporting source-specific complementary information. Appendix~\ref{app:randomization} reports the full statistics. Section~\ref{sec:controlled-coverage} tests whether these sources improve candidate coverage within a shared budget.

\subsection{From Stopping Depth to Candidate Sources}
The preliminary results suggest a different view of recurrent TTS. If deeper computation does not subsume shallower solutions, inference is not only a stopping problem. Under a finite budget, the system must choose depth-specific sources, acquisition counts, and when to stop. This motivates the allocation formulation in the next section.

\takeawaybox{\textbf{Depth is a candidate source as well as a stopping position.} Along a gradually stabilizing latent reasoning trajectory, intermediate states can be decoded again and used as answer sources. Complementary success sets therefore motivate allocating a shared budget across depths.}

\section{Depth Portfolio Allocation}
\label{sec:dpa-method}

\subsection{Budgeted Allocation across Depths}
\label{sec:problem-formulation}
\label{sec:dpa-definition}

The preliminary analysis suggests that a single stopping point can miss correct solutions available at other depths. We therefore formulate \emph{Depth Portfolio Allocation} (DPA): choose candidate sources and acquisition counts under a shared budget to maximize final selection accuracy.

Given a question $q$ and candidate depths $\mathcal{D}$, let $x_{qdr}$ be candidate $r$ at depth $d$, and let $\mathbf{n}=(n_d)_{d\in\mathcal{D}}$ denote the number of candidates acquired from each depth. With nominal per-candidate cost $c_d=d$, the portfolio cost is
\begin{equation}
C(\mathbf{n})=\sum_{d\in\mathcal{D}} d\,n_d \le B,
\label{eq:allocation-budget}
\end{equation}
where $B$ is the budget cap, with at most five candidates per depth in sequential experiments. In our implementation, candidate generation takes tens of seconds, while PPO decisions and selection take milliseconds. Although generated output length affects latency, recurrent computation scales approximately linearly with depth under comparable generation settings; we therefore adopt a nominal recurrent-step budget, excluding decoding and lightweight decision overhead.

Let $R_\alpha$ be the candidate set produced by allocation rule $\alpha$, and let $g(q,R_\alpha)$ select its final answer. DPA seeks
\begin{equation}
\begin{aligned}
\alpha^*=\arg\max_{\alpha\in\mathcal{F}_{\mathrm{DPA}}}
&\;\mathbb{E}\!\left[y\!\left(q,g(q,R_\alpha)\right)\right]\\
\text{s.t.}\quad &C(\mathbf{n}_\alpha)\le B.
\end{aligned}
\label{eq:dpa-objective}
\end{equation}
Final selection accuracy evaluates the returned answer. We separately use oracle coverage---whether a portfolio contains at least one correct candidate---to diagnose source complementarity. DPA therefore covers both predetermined portfolios and sequential acquisition rules.

\begin{figure*}[!t]
\centering
\includegraphics[width=0.98\textwidth]{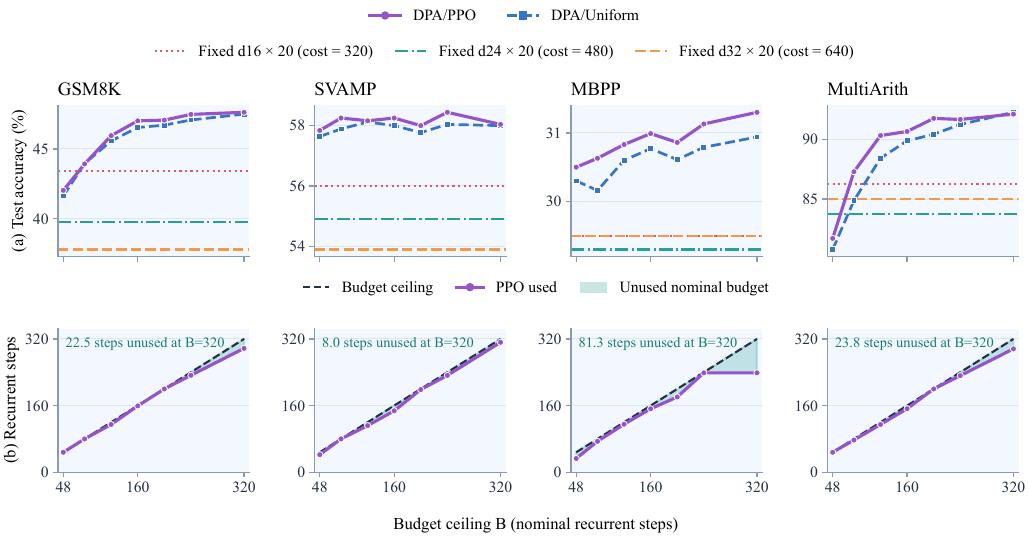}
\caption{\textbf{Accuracy and budget use under shared budget caps.} Top: test accuracy for feedback-guided DPA (PPO) and uniform allocation, with static fixed-depth references. Bottom: PPO nominal cost $C$ versus cap $B$; shading shows unused budget $B-C$, with unused steps annotated at $B=320$. Detailed same-cap comparisons are discussed in Section~\ref{sec:overall-performance}.}
\label{fig:budget-curves}
\vspace{-5mm}
\end{figure*}

\subsection{Feedback-Guided Sequential Allocation}
\label{sec:ppo-depth-selection}

For sequential DPA, the policy observes the question, candidates acquired so far, per-depth counts, selector feedback, and remaining budget. We write the state as
\begin{equation}
s_t=\left(\phi(q),\psi(R_t),\mathbf{n}_t,B-C_t\right),
\label{eq:ppo-state}
\end{equation}
where $\phi(q)$ represents the question and $\psi(R_t)$ summarizes the acquired candidates and selector feedback. At each step the policy chooses
\begin{equation}
a_t\in\mathcal{D}\cup\{\mathrm{stop}\}.
\label{eq:ppo-action}
\end{equation}
If $a_t=d$, one new candidate is acquired at that depth and the portfolio is updated as
\begin{equation}
\begin{aligned}
R_{t+1}&=R_t\cup\{x_{q,d,n_{t,d}+1}\},\\
n_{t+1,d}&=n_{t,d}+1,\qquad C_{t+1}=C_t+d.
\end{aligned}
\label{eq:ppo-update}
\end{equation}
A feasibility mask removes depths that exceed the remaining budget or the per-depth call limit; the stop action makes no further model call.

We instantiate this allocation interface with standard PPO \citep{schulman2017ppo}. On labeled training questions, the reward is the change in correctness of the selector's current output,
\begin{equation}
r_t=u_{t+1}-u_t,
\label{eq:ppo-reward}
\end{equation}
so the undiscounted trajectory return telescopes to final selection correctness. Standard PPO estimates advantages $\hat A_t$ and uses
\begin{equation}
\rho_t(\theta)=\frac{\pi_\theta(a_t\mid s_t)}{\pi_{\theta_{\mathrm{old}}}(a_t\mid s_t)},
\label{eq:ppo-ratio}
\end{equation}
with the clipped surrogate objective
\begin{equation}
\begin{aligned}
L_{\mathrm{PPO}}(\theta)=\mathbb{E}_t\big[\min(&\rho_t\hat A_t,\\
&\operatorname{clip}(\rho_t,1-\epsilon,1+\epsilon)\hat A_t)\big].
\end{aligned}
\label{eq:ppo-objective}
\end{equation}
Ground-truth labels are used only for training rewards; deployment decisions use candidate feedback and budget state. PPO is therefore the optimizer for feedback-guided allocation, rather than the core DPA formulation.

\paragraph{Implementation.}
Our experiments realize each depth action with an unused candidate from a frozen question--depth pool, which isolates allocation and selection from regeneration variance; an online realization would call the recurrent generator after each depth choice. The policy state combines question features, per-depth acquisition counts, remaining budget, selector-score summaries, and projected terminal hidden states from acquired candidates.

\paragraph{Selection and termination.}
After every acquisition, a five-member Borda selector reranks the current subset and returns updated feedback. Depth actions are masked when they violate the remaining budget or the per-depth call limit, and STOP becomes available after the first acquisition. Candidate slots are never reused. Appendix~\ref{app:implementation} gives the exact feature construction, network, masks, selector, cache simulation, and optimization settings.

\section{Experiments}
\label{sec:experiments}

We organize the evaluation around three research questions:
\begin{itemize}
    \item \textbf{RQ1:} Does depth diversity expand the reachable solution set under controlled candidate counts and costs?
    \item \textbf{RQ2:} Does feedback-guided allocation improve final accuracy over uniform allocation under the same budget cap?
    \item \textbf{RQ3:} When can adaptive depth allocation reduce nominal computation rather than simply exhaust the available budget?
\end{itemize}

\subsection{Experimental Setup}

\label{sec:experimental-setup}

\begin{table*}[!t]

\centering

\scriptsize

\setlength{\tabcolsep}{3pt}

\caption{\textbf{RQ1: Oracle coverage (\%) under controlled allocations.} (A) Four candidates per configuration; the portfolio and $4\!\times\!d_{16}$ both cost 64. (B) Matched nominal budgets. Parentheses report gains (pp) over $4\!\times\!d_{16}$ in A and the stronger fixed-depth baseline in B.}

\label{tab:depth-portfolio-control}

\textbf{A. Fixed candidate count}\par\smallskip

\begin{tabular}
{lrrrr}
\toprule
Allocation & Nominal cost & Candidates & MBPP & MultiArith \\
\midrule
$4\!\times\!d_{8}$ & 32 & 4 & 21.04\% & 82.50\% \\
$4\!\times\!d_{16}$ & 64 & 4 & 31.72\% & 91.25\% \\
$4\!\times\!d_{24}$ & 96 & 4 & 31.72\% & 90.25\% \\
$4\!\times\!d_{32}$ & 128 & 4 & 30.88\% & 82.75\% \\
$d_{8}+d_{12}+d_{20}+d_{24}$ & 64 & 4 & \textbf{36.23\%} (+4.51 pp) & \textbf{93.75\%} (+2.50 pp) \\
\bottomrule
\end{tabular}

\medskip

\textbf{B. Fixed nominal cost}\par\smallskip

\resizebox{\textwidth}{!}{%

\begin{tabular}
{cllrrrrr}
\toprule
Budget & Method & Depth allocation & Candidates & GSM8K & SVAMP & MBPP & MultiArith \\
\midrule
\multirow{3}{*}{48}
& Shallow resampling & $6\!\times\!d_{8}$ & 6 & 25.34\% & 48.58\% & 26.44\% & 87.88\% \\
& Fixed intermediate & $3\!\times\!d_{16}$ & 3 & 33.04\% & 58.05\% & 36.04\% & 85.96\% \\
& Depth portfolio & $d_{14}+d_{16}+d_{18}$ & 3 & \textbf{35.20\%} (+2.16 pp) & \textbf{64.50\%} (+6.45 pp) & \textbf{40.40\%} (+4.36 pp) & \textbf{88.70\%} (+0.82 pp) \\
\midrule
\multirow{3}{*}{80}
& Shallow resampling & $10\!\times\!d_{8}$ & 10 & 31.05\% & 51.60\% & 28.90\% & 94.22\% \\
& Fixed intermediate & $5\!\times\!d_{16}$ & 5 & 36.29\% & 59.19\% & 37.22\% & 92.30\% \\
& Depth portfolio & $d_{12}+d_{14}+d_{16}+d_{18}+d_{20}$ & 5 & \textbf{41.70\%} (+5.41 pp) & \textbf{67.00\%} (+7.81 pp) & \textbf{42.80\%} (+5.58 pp) & \textbf{95.70\%} (+1.48 pp) \\
\midrule
\multirow{3}{*}{144}
& Shallow resampling & $18\!\times\!d_{8}$ & 18 & 39.27\% & 54.49\% & 31.49\% & \textbf{97.85\%} \\
& Fixed intermediate & $9\!\times\!d_{16}$ & 9 & 40.30\% & 60.70\% & 37.90\% & 93.40\% \\
& Depth portfolio & $d_{8}+\cdots+d_{24}$ & 9 & \textbf{49.55\%} (+9.25 pp) & \textbf{68.40\%} (+7.70 pp) & \textbf{43.56\%} (+5.66 pp) & \textbf{97.85\%} (+0.00 pp) \\
\bottomrule
\end{tabular}%

}

\vspace{2pt}

\parbox{0.96\textwidth}{\footnotesize\textit{Note:} Oracle coverage is the fraction of questions with at least one correct candidate. Nominal cost sums candidate recurrent depths.}

\vspace{-3mm}
\end{table*}

We first test whether depth portfolios expand oracle coverage, then evaluate selection accuracy and budget use. Experiments cover GSM8K \citep{cobbe2021training}, SVAMP \citep{patel2021svamp}, MultiArith \citep{roy2015solving}, and MBPP \citep{austin2021program} with a frozen Huginn backbone. Appendix~\ref{app:dataset-splits} gives the splits. Within each comparison, methods share prompts and sampling settings. Controlled experiments fix candidate counts or nominal budgets using Table~\ref{tab:depth-portfolio-control}.

Sequential experiments use $\mathcal{D}=\{16,24,32\}$, at most five \emph{on-demand} candidates at each depth, and caps $B\in\{48,80,120,160,200,240,320\}$. Each DPA/PPO depth action acquires one new candidate before the policy continues or stops using the newly updated candidate feedback. Policies are trained separately for each task. Configurations are selected on validation data, then frozen and evaluated on independent test splits. A Borda selector based on latent reward models (LRMs) returns the final answer. We report selection accuracy, nominal cost used, and budget curves to characterize answer quality and computation usage.

\subsection{Does Depth Diversity Expand the Reachable Solution Set? (RQ1)}
\label{sec:controlled-coverage}

To separate source complementarity from additional sampling or computation, we fix candidate counts or nominal budgets and compare oracle coverage for depth portfolios, a fixed intermediate depth, and repeated shallow sampling (Table~\ref{tab:depth-portfolio-control}).

\textbf{Depth portfolios expand coverage at matched candidate counts and nominal costs.} Table~\ref{tab:depth-portfolio-control}A compares a mixed-depth portfolio with a fixed intermediate depth under both experimental controls. Higher coverage on both tasks supports a benefit from combining sources beyond increasing the number of candidates or computational expenditure. Fixed-depth coverage fluctuates across recurrent depths, showing that deeper computation need not access a broader solution set. The matched-count, matched-cost coverage gain also exceeds the fixed-margin null expectation (Appendix~\ref{app:randomization}), further supporting source complementarity.

\textbf{Combining complementary sources can cover more correct answers than adding shallow samples.} Table~\ref{tab:depth-portfolio-control}B compares allocations at equal nominal budgets. Portfolios and the intermediate-depth baseline use the same candidate count, while shallow sampling permits more candidates. Portfolios still achieve higher coverage on GSM8K, SVAMP, and MBPP, indicating that source complementarity affects the solutions reachable within a budget alongside candidate quantity.

\textbf{Portfolio gains narrow when shallow sampling approaches coverage saturation.} MultiArith differs from the other tasks: as the budget increases, repeated shallow sampling approaches the portfolio's coverage and matches it at the highest nominal budget. The value of depth allocation thus depends on which questions a fixed source still misses. When shallow sampling already covers most correct solutions, other depths have less room to contribute further.

\textbf{Answer to RQ1.} Depth diversity expands the reachable solution set beyond candidate quantity alone, but the gain narrows when a strong fixed source approaches coverage saturation.

\takeawaybox{\textbf{Source diversity contributes beyond candidate quantity.} At matched nominal costs, complementary depths expand coverage, with smaller gains when a strong fixed source approaches saturation.}

\subsection{Does Adaptive Allocation Improve Accuracy under the Same Budget? (RQ2)}

\label{sec:experimental-results}

\label{sec:overall-performance}

Table~\ref{tab:combined-results} gives representative absolute performance, while Figure~\ref{fig:budget-curves} and Appendix~\ref{app:exploratory} provide the direct same-cap budget comparison between feedback-guided DPA and DPA/Uniform. Policies are selected on validation data and subsequently evaluated on independent test splits.

\textbf{Feedback-guided allocation generally improves over Uniform under the same cap.} DPA/PPO performs better than Uniform across most budget settings on all four datasets, with the largest gains appearing on MultiArith. Improvements are smaller on the remaining tasks, with only occasional minor regressions.

\textbf{The absolute gains over single-trajectory inference remain substantial, but they reflect a different inference configuration.} At $B=320$, DPA/PPO reaches 47.62\%, 58.03\%, 31.30\%, and 92.12\% on GSM8K, SVAMP, MBPP, and MultiArith. We therefore use the same-cap Uniform comparison, rather than these single-trajectory baselines, as the main test of adaptive allocation.

\textbf{Answer to RQ2.} Under a shared budget cap, feedback-guided allocation usually matches or improves uniform allocation, with the clearest gains on MultiArith at intermediate budgets.

\begin{table}[!t]
\centering\scriptsize\setlength{\tabcolsep}{2.5pt}
\caption{\textbf{Representative DPA results.} A: single-trajectory and latent-correction baselines; B: feedback-guided DPA test accuracy at representative caps; C: nominal cost used. Figure~\ref{fig:budget-curves} and Appendix~\ref{app:exploratory} give same-cap Uniform comparisons.}
\label{tab:combined-results}
\textbf{A. Single-trajectory and latent-corr accuracy (\%)}\par\smallskip
\resizebox{\columnwidth}{!}{%
\begin{tabular}{lrrrr}\toprule
Method & GSM8K & SVAMP & MBPP & MultiArith \\
\midrule
Base / single & 32.60\% & 51.70\% & 27.80\% & 75.00\% \\
CoE-R correction & 33.00\% & 51.00\% & 27.60\% & 71.25\% \\
CoE-C correction & 32.40\% & 51.60\% & 28.00\% & 72.50\% \\
\bottomrule\end{tabular}}
\smallskip
\textbf{B. DPA/PPO selection accuracy (\%)}\par\smallskip
\resizebox{\columnwidth}{!}{%
\begin{tabular}{lrrrr}
\toprule
Budget cap $B$ & GSM8K & SVAMP & MBPP & MultiArith \\
\midrule
80 & 43.92\% & 58.24\% & 30.63\% & 87.29\% \\
200 & 47.06\% & 57.99\% & 30.86\% & 91.76\% \\
320 & 47.62\% & 58.03\% & 31.30\% & 92.12\% \\
\bottomrule
\end{tabular}
}
\smallskip
\textbf{C. DPA/PPO nominal cost used (recurrent steps)}\par\smallskip
\resizebox{\columnwidth}{!}{%
\begin{tabular}{lrrrr}
\toprule
Budget cap $B$ & GSM8K & SVAMP & MBPP & MultiArith \\
\midrule
80 & 80.0 & 80.0 & 74.3 & 77.4 \\
200 & 200.0 & 198.5 & 180.7 & 200.0 \\
320 & 297.5 & 312.0 & 238.7 & 296.2 \\
\bottomrule
\end{tabular}
}
\end{table}

\subsection{When Does Depth Allocation Save Computation? (RQ3)}

\label{sec:accuracy-curves}

Figure~\ref{fig:budget-curves} also reports mean nominal cost $C=\sum_d d n_d$ and unused budget $B-C$ across the seven budget caps. Static fixed-depth references use twenty independent candidates and therefore are not matched-budget comparisons at every horizontal-axis position. Appendix~\ref{app:exploratory} reports the complete test results and budget utilization.

\textbf{Larger budgets do not guarantee monotonic improvement or lower cost.} SVAMP and MBPP selection accuracy exhibits local fluctuations. At $B=320$, PPO uses less nominal computation than Uniform on GSM8K, MBPP, and MultiArith, but slightly more on SVAMP. At smaller budgets, higher accuracy can also entail greater computational expenditure, so a shared budget cap must be distinguished from equal actual cost.

\textbf{PPO can terminate before exhausting its budget.} At larger budgets, the policy often leaves compute unused, suggesting that further acquisition may not help. This effect is most pronounced on MBPP, where PPO uses only 74.6\% of the $B=320$ budget while slightly improving test accuracy over the $B=240$ setting. At $B=320$, PPO also achieves a better accuracy--cost trade-off than Uniform on several tasks. We note that unused budget may arise either from an explicit STOP decision or from infeasible remaining actions, and nominal cost excludes decoding, scoring, and policy-inference overhead.

\textbf{Static fixed-depth references also show that deeper computation need not yield better final answers.}  Deeper configurations have no consistent advantage across tasks, agreeing with the earlier depth observations. Their full nominal costs appear in the legend.

\textbf{Answer to RQ3.} Adaptive allocation saves nominal computation mainly when the policy can stop acquiring useful candidates before the cap is exhausted; this is most visible at high budget caps, especially on MBPP. At larger budgets, the policy often leaves part of the available computation remaining unused, indicating that additional candidate acquisition is not always beneficial.

\takeawaybox{\textbf{Allocation gains depend on the task and budget.} Feedback-guided allocation usually improves over Uniform under the same cap, while computation savings appear when the policy can stop before exhausting the available budget.}

\section{Conclusion}
Recurrent test-time scaling is not only a question of how deep to run. Our preliminary analysis shows that deeper recurrence can lose correct shallower-depth solutions and that depth-specific success sets are broadly non-nested. Motivated by this finding, we formulate \emph{Depth Portfolio Allocation} (DPA), treating recurrent depths as complementary candidate sources under a shared inference budget. Experiments across four benchmark datasets show that mixed-depth portfolios expand candidate solution coverage, while feedback-guided sequential allocation generally matches or improves uniform allocation and can reduce unnecessary recurrent computation. Overall, DPA reframes recurrent TTS from single-depth stopping decisions to budgeted allocation across complementary depths.

\clearpage
\section*{Limitations}
\label{sec:limitations}

Our study focuses on a controlled recurrent reasoning setting with a frozen Huginn backbone and four reasoning/code-generation benchmarks. Extending the analysis to additional recurrent architectures, model scales, and task domains would further test the generality of the depth-allocation view. Our sequential experiments use a bounded candidate pool and nominal recurrent depth as the primary computation measure, which provides a clean basis for comparing allocation strategies. Future work may consider richer candidate-generation settings and end-to-end runtime costs, including decoding and selector overhead. Finally, the current analysis emphasizes depth complementarity and budget allocation; broader forms of adaptive recurrent inference may provide additional opportunities beyond the formulations studied here.

\section*{Ethics Statement}
This work studies test-time computation strategies using established reasoning and code-generation benchmarks. It does not involve human subjects, private personal data, or real-world decision making about individuals. All experiments are conducted offline using model-generated candidates and benchmark-based evaluation. Ground-truth labels are used only for training rewards and evaluation and are not available to the allocation policy at inference time.

\bibliography{references}

\clearpage

\appendix
\small

\section{Deeper Recurrence Does Not Preserve Correctness: Additional Diagnostics}
\label{app:stability-detail}

This appendix extends the preliminary diagnosis in Section~\ref{sec:understanding-steps} across all four tasks and reports the corresponding statistical details.

\begin{table}[!htbp]
\centering
\scriptsize
\setlength{\tabcolsep}{1.8pt}
\renewcommand{\arraystretch}{1.08}

\caption{\textbf{Answer switches under low updates.}
Rates and confidence intervals are percentages.}
\label{tab:stability-detail}

\begin{tabularx}{0.99\columnwidth}{
@{}
l
c
c
c
c
>{\centering\arraybackslash}X
>{\centering\arraybackslash}X
@{}
}
\toprule
Task &
$\tau_u$ &
$\tau_v$ &
\shortstack{Low-update\\intervals} &
\shortstack{Truncated\\answers} &
\shortstack{Switch\\rate\\{[95\% CI]}} &
\shortstack{Non-\\truncated\\rate\\{[95\% CI]}} \\
\midrule

GSM8K
& 0.07364
& 0.002711
& 323
& 56
& 25.08 [14.33, 37.20]
& 25.82 [15.04, 38.23] \\

SVAMP
& 0.05206
& 0.001355
& 266
& 1
& 9.40 [0.84, 20.20]
& 9.40 [0.84, 20.20] \\

MultiArith
& 0.04950
& 0.001225
& 317
& 3
& 21.14 [11.91, 31.07]
& 20.89 [11.65, 30.79] \\

MBPP
& 0.03029
& 0.000459
& 288
& 73
& 9.72 [3.91, 16.39]
& 10.33 [4.25, 17.31] \\

\bottomrule
\end{tabularx}
\end{table}

\begin{table}[H]
\centering
\scriptsize
\setlength{\tabcolsep}{3pt}
\renewcommand{\arraystretch}{1.08}
\caption{\textbf{Correctness transitions under low updates.}
All low-update intervals form the denominator; values are percentages.}
\label{tab:transition-detail}

\begin{tabularx}{0.95\columnwidth}{
@{}
l
>{\centering\arraybackslash}X
>{\centering\arraybackslash}X
>{\centering\arraybackslash}X
@{}
}
\toprule
Task &
Wrong$\to$correct &
Correct$\to$wrong &
Wrong$\to$other wrong \\
\midrule
GSM8K
& 1.24 [0.00, 3.13]
& 4.33 [1.57, 7.62]
& 19.50 [9.54, 31.38] \\

SVAMP
& 0.00 [0.00, 0.00]
& 0.00 [0.00, 0.00]
& 9.40 [0.84, 20.20] \\

MultiArith
& 5.05 [1.02, 9.79]
& 5.68 [1.27, 10.84]
& 10.41 [3.54, 18.68] \\

MBPP
& 1.04 [0.00, 3.25]
& 0.35 [0.00, 1.08]
& 3.82 [0.67, 8.24] \\
\bottomrule
\end{tabularx}
\end{table}

\paragraph{Paired diagnostics and statistical units.}
Each task has twenty test questions, five repeats, and thirteen depths: 1,300 complete answers and 1,200 adjacent-depth intervals. States are taken at the final prompt token during recurrent computation. Elementwise checks confirm shared-prefix consistency for all four tasks. Subsequent answers are generated at their respective fixed depths. Changes therefore reflect the complete depth-specific generation procedure, without isolating the causal effect of small prompt-boundary state changes.

\paragraph{Update metrics and thresholds.}
For prompt-boundary state $h_d$, define $u_d=\|h_d-h_{d-2}\|_2/(\|h_{d-2}\|_2+10^{-12})$ and $v_d=1-\cos(h_d,h_{d-2})$. Low-update intervals satisfy both $u_d\le\tau_u$ and $v_d\le\tau_v$. Each threshold is the 25th percentile of its metric on independent training questions (Table~\ref{tab:stability-detail}). This condition denotes relatively small updates rather than mathematical convergence.

\paragraph{Intervals and truncation sensitivity.}
We bootstrap questions with replacement 10,000 times using seed 20260922, retaining within-question depths and repeats. Curves average repeats within each question before averaging questions. Conditional event rates divide event counts by the number of qualifying intervals. The non-truncated analysis requires both adjacent answers to terminate below the generation cap; Table~\ref{tab:stability-detail} reports its switch rates separately.

\begin{figure*}[!htbp]
    \centering
    \includegraphics[width=0.89\textwidth]{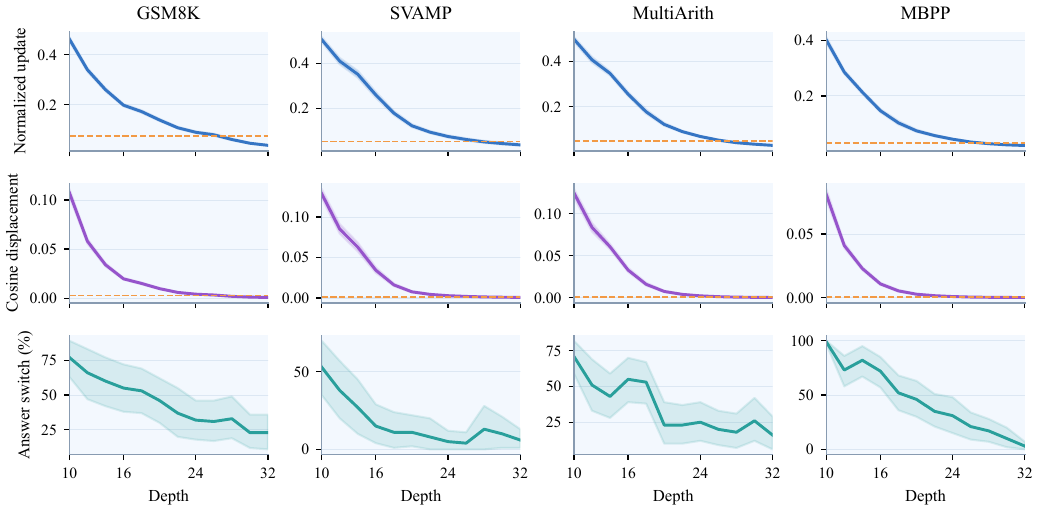}
    \caption{\textbf{State updates and answer switches.} Shading shows question-bootstrap 95\% intervals; dashed lines indicate training-calibrated thresholds.}
    \label{fig:stability-four}
\end{figure*}

\paragraph{Interpretation.}
\textbf{Answers can switch under low updates, with task-dependent implications.} Figure~\ref{fig:stability-four} extends the main-text example. Table~\ref{tab:transition-detail} separates wrong-to-correct, correct-to-wrong, and wrong-to-different-wrong events. No transitions of the first two types are observed in SVAMP's low-update intervals; other tasks exhibit both repairs and degradations. Two correct programs can also differ in identity, so the three correctness events need not exhaust code switches. Representation stability, answer-identity stability, and correctness stability should therefore be measured separately.

\section{Correct Solutions Are Complementary across Depths: Additional Statistics}
\label{app:depth-detail}

This appendix complements Section~\ref{sec:complementary-depth-solutions} with full per-depth statistics and directional rescue results.

\begin{figure*}[t]
    \centering
    \includegraphics[width=\textwidth]{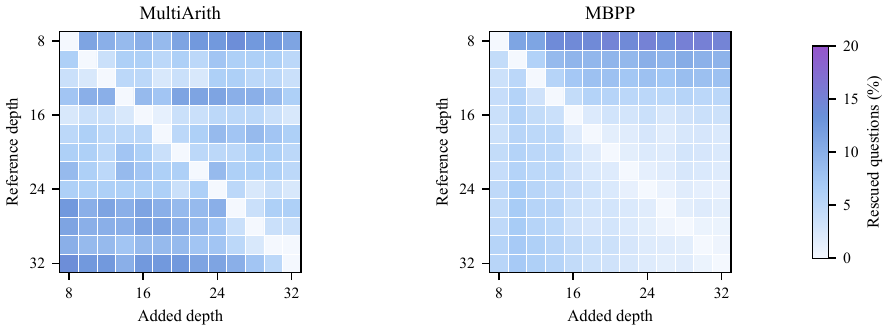}
    \caption{\textbf{Directional rescue across depths.} Rows denote the reference depth and columns the added depth. Color shows the fraction of questions solved only by the added depth.}
    \label{fig:directional-rescue}
\end{figure*}

\begin{table*}[t]
\centering
\scriptsize
\setlength{\tabcolsep}{7pt}
\caption{\textbf{Full results at thirteen depths.} MA: 80-question MultiArith confirmatory pool; MBPP: 500-question grid pool.}
\label{tab:depth-full}
\begin{tabular}{rrrrrrr}
\toprule
$d$ & \shortstack{MA single\\(\%)} & \shortstack{MA solved\\questions} & \shortstack{MA Oracle@5\\(\%)} & \shortstack{MBPP single\\(\%)} & \shortstack{MBPP solved\\questions} & \shortstack{MBPP Oracle@5\\(\%)} \\
\midrule
8  & 50.25 & 69 & 86.25 & 15.16 & 109 & 21.80 \\
10 & 64.00 & 73 & 91.25 & 22.24 & 143 & 28.60 \\
12 & 66.00 & 74 & 92.50 & 24.12 & 147 & 29.40 \\
14 & 69.25 & 70 & 87.50 & 26.36 & 158 & 31.60 \\
16 & 71.75 & 75 & 93.75 & 27.96 & 161 & 32.20 \\
18 & 74.25 & 72 & 90.00 & 29.04 & 162 & 32.40 \\
20 & 73.00 & 73 & 91.25 & 29.52 & 161 & 32.20 \\
22 & 70.75 & 72 & 90.00 & 28.96 & 159 & 31.80 \\
24 & 68.50 & 74 & 92.50 & 29.48 & 160 & 32.00 \\
26 & 70.50 & 70 & 87.50 & 28.52 & 156 & 31.20 \\
28 & 71.75 & 70 & 87.50 & 28.92 & 161 & 32.20 \\
30 & 71.50 & 71 & 88.75 & 29.12 & 157 & 31.40 \\
32 & 70.50 & 67 & 83.75 & 29.00 & 156 & 31.20 \\
\bottomrule
\end{tabular}
\end{table*}

\paragraph{Marginal performance and coverage.}
Table~\ref{tab:depth-full} recomputes three metrics from raw correctness tensors. Single-sample accuracy averages five repeats per question; solved counts indicate at least one correct repeat; Oracle@5 divides these counts by pool size. Correlations among repeats remain in the observed pool: Oracle@5 is computed directly from its five candidates rather than approximated by independent Bernoulli draws. Per-depth results are also saved as CSV for checking Figure~\ref{fig:success-set-jaccard} and the success-set definition.

\paragraph{Directional rescue.}
Jaccard distance summarizes the overall difference between two depth-specific success sets, but it is symmetric and therefore does not reveal which depth contributes the additional solved questions. Figure~\ref{fig:directional-rescue} complements this analysis with a directional measure. For a reference depth $d_i$ and an added depth $d_j$, we define
\begin{equation}
R_{i\to j}=\frac{|C_{d_j}\setminus C_{d_i}|}{Q},
\end{equation}
where $Q$ is the total number of questions in the evaluation pool. The numerator counts questions that are solved by at least one of the five candidates from $d_j$ but by none of the five candidates from $d_i$. Thus, $R_{i\to j}$ measures the fraction of the entire pool uniquely rescued by adding source $d_j$ to source $d_i$. Because the measure is directional, $R_{i\to j}$ and $R_{j\to i}$ need not be equal: one depth may recover many questions missed by another while contributing little in the reverse direction. Diagonal entries are zero by definition. Each cell therefore characterizes source-level complementarity between two complete depth-specific candidate sets, rather than the marginal gain of acquiring one additional candidate under a matched computation budget.

\paragraph{Connection to best-depth preferences.}
Figure~\ref{fig:step-preference-diversity} identifies, for each question, the depth with the highest five-repeat accuracy, with tied maxima split equally and unsolved questions excluded. Preference frequencies indicate which depths are most often favored, Oracle@5 measures per-depth coverage, and directional rescue captures failures of one depth recovered by another. Together, these metrics distinguish depth preference, overall coverage, and complementary coverage. A depth can therefore perform well on average while still missing questions solved elsewhere, motivating portfolio allocation across depths.

\paragraph{Statistical interpretation.}
\textbf{Marginal performance and directional rescue jointly characterize depth value.} Table~\ref{tab:depth-full} shows non-monotonic accuracy and solved counts, while the heatmap identifies omissions recovered by an added source. These statements apply within each pool. The early MultiArith preference pool and confirmatory set pool are analyzed separately; cross-pool differences are not interpreted as depth effects, and code-text disagreement on MBPP is not equated with functional disagreement.

\section{Feedback-Guided Sequential Allocation: Implementation Details}
\label{app:implementation}

This appendix provides the implementation details for the feedback-guided allocation rule in Section~\ref{sec:ppo-depth-selection}. The sequential state, portfolio update, reward, PPO ratio, and clipped objective are defined in the main text and are not repeated here.

\paragraph{Sequential execution and cache simulation.}
DPA's decision interface selects a depth or stops within a budget. The cache implementation simulates this interface with frozen candidates: after a depth action, it uniformly draws an unused slot from that depth's five candidates, then updates feedback and cost. It does not regenerate answers or update Huginn or the LRMs. It reproduces allocation and selection on the given pool; online execution instead calls the generator after each depth action to produce one candidate. The parameters below describe this cache simulation.

\paragraph{State features.}
Question text is lowercased and split into words, numbers, and symbols. Signed hashing of unigrams and adjacent bigrams yields an $\ell_2$-normalized 256-dimensional vector. The five LRM scores for acquired candidates are standardized per LRM over the current subset, then summarized by per-depth means and maxima. Additional features are counts divided by five, the remaining-budget fraction, and the initial budget divided by 320. Unacquired terminal states are zero-filled; acquired $h_T$ vectors pass through LayerNorm and a 32-dimensional projection, then are averaged by depth. Correctness labels are used only for training rewards and offline evaluation.

\paragraph{Network.}
Text, score, and budget branches project to 96, 96, and 32 dimensions with Tanh. The hidden-state branch retains four depth slots of 32 dimensions each; the $d=8$ slot remains empty in this analysis. Concatenated features pass through a two-layer Tanh body with widths $192\to128$, producing action logits and a scalar value. Five action slots represent four depths and stopping. Masking disables $d=8$, leaving $16,24,32$, consistent with the three-source formulation in the main text.

\paragraph{Masks and termination.}
A depth is feasible only if it fits the remaining budget and has been used fewer than five times. The cache implementation forbids stopping with an empty initial set and permits it after the first acquisition. If no depth remains feasible, it terminates and returns the current selector output. Costs, counts, and masks are updated after every acquisition. Slots are never reused, at most fifteen candidates can be acquired per question, and unused budget may remain.

\paragraph{Five-member Borda selector.}
Each frozen LRM ranks only the current subset, assigning higher ranks to higher scores. The candidate with the highest mean rank across five members is selected. Stable sorting resolves ties within a member; ties in final mean rank return the first candidate in the current index order. Ranks are recomputed whenever the subset changes. Standardized scores enter the policy state, whereas subset ranks determine the final answer.

\begin{table}[!htbp]
\centering
\footnotesize
\setlength{\tabcolsep}{4pt}
\caption{\textbf{Optimization settings for the PPO implementation.}}
\label{tab:ppo-settings}
\begin{tabular}{lr}
\toprule
Parameter & Setting \\
\midrule
Parallel environments / horizon & 24 / 22 \\
Epochs per update / minibatch & 4 / 128 \\
Discount $\gamma$ / GAE $\lambda$ & 1 / 0.95 \\
PPO clip $\epsilon$ & 0.2 \\
Value coefficient / max. gradient norm & 0.5 / 0.5 \\
Optimizer & Adam \\
Updates (script default) & 160 \\
Selection / evaluation draws (default) & 12 / 48 \\
\bottomrule
\end{tabular}
\end{table}

\paragraph{Reward and optimization.}
Rewards are changes in frozen-selector correctness after acquisition, with no additional cost penalty. Stopping acquires no candidate. Advantages are computed per question trajectory and standardized within the update batch; optimization combines the clipped policy loss, value loss, and entropy term. Table~\ref{tab:ppo-settings} lists the settings. Update counts and evaluation draws are command-line-overridable defaults. The included historical source preserves their definitions; training reproduction should also record the actual command and checkpoint configuration.

\begin{table}[t]
\centering
\footnotesize
\setlength{\tabcolsep}{5pt}
\caption{\textbf{Historical tuning configurations.}}
\label{tab:ppo-trials}
\begin{tabular}{rrrr}
\toprule
Trial index & Learning rate & Entropy coefficient & Seed \\
\midrule
0 & 1e-04 & 0.001 & 13 \\
1 & 1e-04 & 0.001 & 29 \\
2 & 1e-04 & 0.01 & 13 \\
3 & 1e-04 & 0.01 & 29 \\
4 & 3e-04 & 0.001 & 13 \\
5 & 3e-04 & 0.001 & 29 \\
6 & 3e-04 & 0.01 & 13 \\
7 & 3e-04 & 0.01 & 29 \\
\bottomrule
\end{tabular}
\end{table}

\paragraph{Configuration selection.}
At each budget, configurations are selected independently by validation accuracy; ties favor lower cost, then the smaller trial index. Selected policies are frozen and evaluated on independent test splits. All eight configurations in Table~\ref{tab:ppo-trials} use acquired candidates' $h_T$.

\FloatBarrier

\section{Experimental Setup: Data and Evaluation Protocols}
\label{app:dataset-splits}

This appendix expands the experimental setup in Section~\ref{sec:experimental-setup} and records dataset splits, analysis pools, matching rules, and evaluation details.

\paragraph{Dataset splits.}
Table~\ref{tab:dataset-splits} lists the dataset splits. Math tasks use questions and final numerical answers; MBPP uses programming tasks and test cases. Training data fit allocation policies, validation data select configurations, and independent test splits provide the results in Table~\ref{tab:combined-results}, Figure~\ref{fig:budget-curves}, and Appendix~\ref{app:adaptive-full}. Validation and test data are kept separate; selected configurations are frozen before test evaluation.

\begin{table}[t]
\centering
\footnotesize
\setlength{\tabcolsep}{8pt}
\caption{\textbf{Dataset splits.}}
\label{tab:dataset-splits}
\begin{tabular}{lrrr}
\toprule
Dataset & Train & Validation & Test \\
\midrule
GSM8K & 6,473 & 500 & 1,319 \\
SVAMP & 700 & 150 & 150 \\
MBPP & 369 & 90 & 496 \\
MultiArith & 330 & 90 & 80 \\
\bottomrule
\end{tabular}
\end{table}

\paragraph{Analysis pools.}
Table~\ref{tab:pool-protocols} gives the scope of each analysis. State diagnostics compare recurrent states at the shared prompt boundary and subsequent answers on twenty test questions per task. Fine-grid analyses use thirteen depths and five repeats per depth. MultiArith depth preferences in Figure~\ref{fig:step-preference-diversity} use an early 180-question pool; Figure~\ref{fig:success-set-jaccard} and Appendix~\ref{app:depth-detail} use an 80-question confirmatory pool. Their question scopes and prompt batches differ, so question-level results are kept separate. The 500-question MBPP grid pool is also reported separately from the sequential test split.

\begin{table}[t]
\centering
\scriptsize
\setlength{\tabcolsep}{3pt}
\caption{\textbf{Analysis pools and purposes.}}
\label{tab:pool-protocols}
\begin{tabular}{lll}
\toprule
Analysis & Questions & Depths / repeats \\
\midrule
State diagnostics & 20 per task & $8{:}2{:}32$ / 5 each \\
Depth preferences & MA 180 / MBPP 500 & $8{:}2{:}32$ / 5 each \\
Success sets & MA 80 / MBPP 500 & $8{:}2{:}32$ / 5 each \\
Budget evaluation & Independent test splits & $16,24,32$ / up to 5 \\
\bottomrule
\end{tabular}
\end{table}

\paragraph{Matching inputs, candidates, and labels.}
Each question uses a fixed prompt template. Depth comparisons within an analysis pool share prompts, questions, and scoring rules. A question identifier, depth, and repeat index uniquely identify each candidate. The fine-grid correctness tensor is $Y\in\{0,1\}^{Q\times13\times5}$, with depths $8,10,\ldots,32$ and five generations. Sequential caches retain correctness, frozen LRM scores, and terminal hidden states. These fields are joined by question identifier while preserving depth and repeat axes, ensuring that feedback and labels refer to the same candidates.

\paragraph{Diagnostic prompts and generation.}
Diagnostics use Huginn-0125 with bfloat16 and greedy fixed-depth decoding: \texttt{do\_sample=False}, at most 512 new tokens, and a total length cap of 4096. Math prompts contain eight examples organized with \texttt{Question:} and \texttt{Answer:}; the full template is included in the diagnostic prompt file. The project's code-task prompt builder supplies MBPP examples and task descriptions. Seeds derived from the question, repeat index, and token position pair shared-prefix states across depths. Repeats mainly reflect stochastic recurrent states. These generation settings apply specifically to this diagnostic batch.

\paragraph{Diagnostic correctness and answer identity.}
Math evaluation extracts the number following the last \texttt{\#\#\#\#}, or otherwise the final number; absolute error below 0.01 is correct. Code evaluation removes special tokens and executes task tests with a 5-second CPU limit, 2-GiB memory limit, and 7-second subprocess timeout. Math identity uses normalized numbers. Code identity uses an abstract syntax tree without location attributes, falling back to normalized text if parsing fails. Answer switches can therefore change correctness or occur between answers that are both correct or both wrong.

\paragraph{Metrics and cost.}
Oracle coverage is the fraction of questions with at least one correct candidate; selection accuracy is the fraction for which the selector returns a correct candidate. They share correctness labels but evaluate a set and a returned answer, respectively. Nominal cost sums acquired candidate depths, $C=\sum_d d n_d$, with cap $B$. It excludes decoding length, selector runtime, and policy inference. Split sizes, pool sizes, and repeats are reported within their respective protocols.

\section{Does Depth Diversity Expand the Reachable Solution Set? Additional Tests}
\label{app:randomization}
\label{app:fixed-20}

This appendix reports the fixed-margin tests and static fixed-depth references used to support RQ1 in Section~\ref{sec:controlled-coverage}.

\paragraph{Fixed-margin null.}
The correctness tensor is flattened into a $Q\times65$ binary matrix, with questions as rows and depth--repeat slots as columns. Randomization fixes both margins, preserving total successes per question and slot. We run 5,000 randomizations per dataset across five chains, reconstructing success sets and statistics to test whether depth structure exceeds these marginal constraints.

\paragraph{Statistics and correction.}
Mutual non-nesting counts all 78 depth pairs. The Jaccard slope uses the controlled regression in the main text and is scaled per eight steps. Oracle gains compare the specified matched-cost configurations. One-sided $p$ values use the fraction of null statistics at least as large as observed, with a finite-replicate correction. Holm correction jointly covers six tests.

\paragraph{Non-nesting and randomization.}
MBPP's count of non-nested depth pairs does not significantly exceed its null expectation, whereas its adjusted slope and matched-cost gain do. Observed set relationships and fixed-margin tests characterize structure and its statistical interpretation separately; significance for one statistic does not imply significance for another.

\paragraph{Static fixed-depth references.}
Table~\ref{tab:fixed-20} gives static references with twenty candidates per fixed depth, costing 320, 480, and 640, respectively. They form the horizontal lines in Figure~\ref{fig:budget-curves}. These configurations are reported separately from sequential pools with at most five candidates per depth; sequential caps do not represent the static configurations' actual costs.

\begin{table}[!htbp]
\centering
\footnotesize
\setlength{\tabcolsep}{2.5pt}
\renewcommand{\arraystretch}{1.08}

\caption{\textbf{Accuracy and nominal budgets for twenty candidates at each fixed depth.}}
\label{tab:fixed-20}

\begin{tabularx}{0.99\columnwidth}{
@{}
>{\raggedright\arraybackslash}X
c
c
c
c
c
@{}
}
\toprule
Method &
$B$ &
GSM8K &
SVAMP &
MBPP &
MultiArith \\
\midrule

Fixed $d_{16}\!\times\!20$
& 320
& 43.40\%
& 56.00\%
& 29.50\%
& 86.25\% \\

Fixed $d_{24}\!\times\!20$
& 480
& 39.80\%
& 54.90\%
& 29.30\%
& 83.75\% \\

Fixed $d_{32}\!\times\!20$
& 640
& 37.80\%
& 53.90\%
& 29.50\%
& 85.00\% \\

\bottomrule
\end{tabularx}
\end{table}

\section{Does Adaptive Allocation Improve Accuracy under the Same Budget? Complete Results}
\label{app:adaptive-full}
\label{app:exploratory}

This appendix provides the complete same-cap comparison used for RQ2 in Section~\ref{sec:experimental-results}.

\begin{table}[H]
\centering
\scriptsize
\setlength{\tabcolsep}{1.5pt}
\renewcommand{\arraystretch}{1.05}

\caption{\textbf{Complete test-budget results.}
Costs are mean recurrent steps; differences are PPO minus Uniform.
Values are rounded independently.}
\label{tab:budget-full}

\begin{tabularx}{0.99\columnwidth}{
@{}
l
c
>{\centering\arraybackslash}X
>{\centering\arraybackslash}X
>{\centering\arraybackslash}X
>{\centering\arraybackslash}X
>{\centering\arraybackslash}X
@{}
}
\toprule
Task &
$B$ &
\shortstack{PPO\\(\%)} &
\shortstack{PPO\\cost} &
\shortstack{Uniform\\(\%)} &
\shortstack{Uniform\\cost} &
\shortstack{$\Delta$\\(pp)} \\
\midrule

\multirow{7}{*}{GSM8K}
& 48  & 42.03 & 48.0  & 41.63 & 40.0  & +0.41 \\
& 80  & 43.92 & 80.0  & 43.96 & 72.0  & -0.04 \\
& 120 & 45.95 & 114.7 & 45.57 & 112.0 & +0.38 \\
& 160 & 47.01 & 159.3 & 46.53 & 160.0 & +0.48 \\
& 200 & 47.06 & 200.0 & 46.70 & 184.0 & +0.35 \\
& 240 & 47.47 & 232.7 & 47.07 & 232.0 & +0.39 \\
& 320 & 47.62 & 297.5 & 47.49 & 304.0 & +0.12 \\
\midrule

\multirow{7}{*}{SVAMP}
& 48  & 57.83 & 42.3  & 57.63 & 40.0  & +0.20 \\
& 80  & 58.24 & 80.0  & 57.88 & 72.0  & +0.36 \\
& 120 & 58.15 & 112.0 & 58.11 & 112.0 & +0.04 \\
& 160 & 58.24 & 147.5 & 57.99 & 160.0 & +0.26 \\
& 200 & 57.99 & 198.5 & 57.76 & 184.0 & +0.24 \\
& 240 & 58.43 & 232.5 & 58.03 & 232.0 & +0.40 \\
& 320 & 58.03 & 312.0 & 57.99 & 304.0 & +0.05 \\
\midrule

\multirow{7}{*}{MBPP}
& 48  & 30.50 & 33.1  & 30.30 & 40.0  & +0.20 \\
& 80  & 30.63 & 74.3  & 30.16 & 72.0  & +0.47 \\
& 120 & 30.83 & 115.3 & 30.60 & 112.0 & +0.24 \\
& 160 & 30.99 & 152.4 & 30.77 & 160.0 & +0.22 \\
& 200 & 30.86 & 180.7 & 30.61 & 184.0 & +0.26 \\
& 240 & 31.13 & 238.7 & 30.79 & 232.0 & +0.33 \\
& 320 & 31.30 & 238.7 & 30.94 & 304.0 & +0.36 \\
\midrule

\multirow{7}{*}{MultiArith}
& 48  & 81.70 & 48.0  & 80.77 & 40.0  & +0.93 \\
& 80  & 87.29 & 77.4  & 84.87 & 72.0  & +2.43 \\
& 120 & 90.32 & 114.7 & 88.42 & 112.0 & +1.90 \\
& 160 & 90.66 & 152.6 & 89.88 & 160.0 & +0.78 \\
& 200 & 91.76 & 200.0 & 90.43 & 184.0 & +1.34 \\
& 240 & 91.67 & 231.9 & 91.25 & 232.0 & +0.42 \\
& 320 & 92.12 & 296.2 & 92.27 & 304.0 & -0.15 \\
\bottomrule
\end{tabularx}
\end{table}

\paragraph{Evaluation and configurations.}
These results share the source of Figure~\ref{fig:budget-curves}. Policies are fitted on training data, configurations are selected on validation data, and accuracy and mean cost are evaluated on independent test splits. The seven caps range from 48 to 320. Accuracy, cost, and PPO-minus-Uniform differences are rounded independently, so a reported difference can differ by 0.01 percentage points from subtracting the two displayed accuracies.

\paragraph{Statistical interpretation.}
Budget comparisons report point estimates of selection accuracy, mean nominal cost, and PPO-minus-Uniform differences. These point estimates do not establish statistical significance. A shared budget cap does not imply equal actual expenditure; allocation strategies should be compared in terms of both accuracy and cost.

\FloatBarrier

\section{When Does Depth Allocation Save Computation? Budget Utilization}
\label{app:budget-utilization}

This appendix reports the budget-utilization details used for RQ3 in Section~\ref{sec:accuracy-curves} and makes the Uniform baseline explicit.

\begin{table*}[t]
\centering
\footnotesize
\setlength{\tabcolsep}{8pt}
\caption{\textbf{PPO budget use on independent test splits.} Cells report mean nominal recurrent steps $C$, with budget utilization $100C/B$ in parentheses. Cap $B$ is the maximum permitted nominal cost, not runtime.}
\label{tab:budget-utilization}
\begin{tabular}{rrrrr}
\toprule
$B$ & GSM8K & SVAMP & MBPP & MultiArith \\
\midrule
48  & 48.0 (100.0\%) & 42.3 (88.1\%) & 33.1 (69.0\%) & 48.0 (100.0\%) \\
80  & 80.0 (100.0\%) & 80.0 (100.0\%) & 74.3 (92.9\%) & 77.4 (96.8\%) \\
120 & 114.7 (95.6\%) & 112.0 (93.3\%) & 115.3 (96.1\%) & 114.7 (95.6\%) \\
160 & 159.3 (99.6\%) & 147.5 (92.2\%) & 152.4 (95.2\%) & 152.6 (95.4\%) \\
200 & 200.0 (100.0\%) & 198.5 (99.2\%) & 180.7 (90.3\%) & 200.0 (100.0\%) \\
240 & 232.7 (97.0\%) & 232.5 (96.9\%) & 238.7 (99.5\%) & 231.9 (96.6\%) \\
320 & 297.5 (93.0\%) & 312.0 (97.5\%) & 238.7 (74.6\%) & 296.2 (92.6\%) \\
\bottomrule
\end{tabular}
\end{table*}

\paragraph{Exact uniform-allocation rule.}
At each step, Uniform considers the least-used depths, choosing the shallowest affordable one among them. If none is affordable, it stops rather than acquiring from a more-used source. Table~\ref{tab:uniform-allocation} gives the shared allocations and costs across tasks, explaining the Uniform costs in Table~\ref{tab:budget-full}.

\begin{table}[t]
\centering
\footnotesize
\setlength{\tabcolsep}{5pt}
\caption{\textbf{Uniform allocations in budget evaluation.}}
\label{tab:uniform-allocation}
\begin{tabular}{rrrrr}
\toprule
$B$ & $n_{16}$ & $n_{24}$ & $n_{32}$ & Nominal cost \\
\midrule
48  & 1 & 1 & 0 & 40 \\
80  & 1 & 1 & 1 & 72 \\
120 & 2 & 2 & 1 & 112 \\
160 & 3 & 2 & 2 & 160 \\
200 & 3 & 3 & 2 & 184 \\
240 & 4 & 3 & 3 & 232 \\
320 & 5 & 4 & 4 & 304 \\
\bottomrule
\end{tabular}
\end{table}

\FloatBarrier

\section{Reproduction Notes}
\label{app:reproduction}

\paragraph{Reproduction scope.}
The submission source compiles directly from \texttt{full\_naacl.tex}; all tables are embedded in that file and the reported figures are supplied as PDF assets. Full regeneration of the figures and retraining additionally require the frozen diagnostic tensors, depth grids, PPO summaries, prompt examples, model and LRM checkpoints, launch commands, and software environment, which are maintained separately from this manuscript-only archive.

\paragraph{Data checks.}
Candidate-level source records for Figure~\ref{fig:latent-convergence}(b)'s original example have not been located. Some original controlled-table MultiArith values differ slightly from later recomputed summaries; the original controlled values are retained. Diagnostic, confirmatory-grid, and test-budget summaries are analyzed under their respective protocols.

\end{document}